\documentclass[sigconf,nonacm]{acmart}
\setcopyright{none}

\usepackage{xurl} 
\usepackage{enumitem}
\usepackage{amsmath}
\usepackage{array}
\usepackage{makecell}
\usepackage{multirow}
\usepackage{booktabs}
\usepackage{subcaption}
\usepackage[normalem]{ulem}
\usepackage{graphicx}
\usepackage{tabularx}
\usepackage{amsmath}

\makeatletter
\newcommand\workshophead{``Nobody Did This'' Workshop at CSCW '26,
  October 10--14, 2026, Salt Lake City, UT, USA}
\AtBeginDocument{%
  \let\ws@standardpagestyle\ps@standardpagestyle
  \def\ps@standardpagestyle{\ws@standardpagestyle
    \fancyhead[LE]{\ACM@linecountL\@headfootfont\workshophead}%
    \fancyhead[RO]{\@headfootfont\workshophead\ACM@linecountR}}%
  \pagestyle{standardpagestyle}}
\makeatother

\AtBeginDocument{%
  }

\setcopyright{acmlicensed}
\copyrightyear{2026}
\acmYear{2026}
\acmDOI{}
\acmISBN{}

\acmConference[CSCW '26 Workshop]{the 29th ACM conference on computer-supported cooperative work and social computing}{October 10--14,
  2026}{Salt Lake City, Utah, USA}

\begin{document}

\title[When No One Owns the Judgment]{When No One Owns the Judgment: Accountability Under Contribution Dissolution in Human–AI Collaboration}\titlenote{Position paper accepted to the workshop ``Nobody Did This'':
Contribution, Originality, and Accountability in Agent-Mediated
Collaboration~\cite{Imteyaz2026-nobody} at the 2026 ACM Conference on
Computer-Supported Cooperative Work and Social Computing (CSCW '26),
Salt Lake City, UT, USA, October 2026.}

\author{Hengzhi Ye}
\affiliation{%
  \institution{School of Computer Science}
  \institution{Peking university}
  \city{Beijing}
  \country{China}
}
\email{hzye@stu.pku.edu.cn}


\renewcommand{\shortauthors}{Hengzhi Ye}

\begin{abstract}
Communities often respond to potentially AI-assisted work by asking three questions: Was AI used? Was that use disclosed? Can hidden use be detected? These questions place AI use itself at the center of accountability while overlooking a deeper problem: unowned judgment. Evaluations, claims, decisions, and creative directions can be shaped by AI with no accountable human or institution prepared to stand behind them. We develop this argument through two illustrative cases: AI-assisted peer review and concealed AI use in creative work. The first shows how contribution dissolution can weaken responsibility while the second shows how the fear of losing credit can discourage honest disclosure. The cases expose the limits of disclosure rules and provenance records as responses to AI-mediated collaboration. We offer three directions for discussion: distinguishing the roles AI plays, identifying judgments that require clear human ownership, and creating conditions in which AI involvement can be disclosed without default penalty. The broader aim is to make AI-shaped contributions discussable, creditable, contestable, and repairable.

\end{abstract}


\maketitle

\section{Introduction: Beyond “Did You Use AI?”}

Human-AI collaboration rarely follows a clean separation of contribution. A person may write every sentence while the system shapes the central angle, constraints, or evaluative criteria; a system may generate most of the text while simply carrying out a plan already formed by the human. The final outcome therefore reveals little about how the underlying judgments emerged. This reflects a long-standing concern in HCI that human and machine agency is configured through situated interaction instead of being divided along a fixed boundary \cite{suchman2007humanmachine}.

Many institutional responses reduce this complex process to a simple question: \textit{Did you use AI?} The question fits easily into a disclosure form, a policy statement, or a detection system, yet it treats different forms of assistance as equivalent. It directs attention to the presence of AI while leaving aside how judgments are formed and who stands behind them. The deeper problem arises when an AI-shaped evaluation, claim, decision, or creative direction enters consequential work without an accountable actor who can answer for its consequences, which we refer to as \emph{unowned judgment}.

The distinction becomes clearer in peer review. A reviewer may use AI to refine the wording of a review they developed independently while retaining full understanding of and responsibility for the underlying judgment. In a different case, a reviewer may submit model-generated objections without being able to evaluate or defend them. Although both practices involve AI, they raise different accountability concerns. A similar distinction applies in journalism: using AI to brainstorm headlines differs substantially from adopting an AI-shaped factual framing that has not been verified by an editor.

Policies centered on AI use struggle to account for these differences. They may penalize limited forms of assistance while failing to specify when a person has delegated a judgment they were expected to make themselves. Research on computerized accountability has long shown that distributed systems can make responsibility difficult to locate \cite{nissenbaum1996accountability}. Work on moral crumple zones further shows how responsibility may fall on the nearest human even when that person had little meaningful control \cite{elish2019moral}. Human--AI collaboration creates a related tension: a person may retain formal responsibility for a judgment whose substance was largely shaped by an AI system.

The core provocation is therefore simple: asking whether AI was used may still matter, but it cannot tell us whether anyone truly owns the resulting judgment. Accountability should begin with the judgments that people and institutions are expected to understand, endorse, and defend. It should also ask who is responsible for correction when those judgments cause harm. 

\section{Related Work: Contribution Attribution}

Contribution attribution becomes difficult when AI shapes more than the final artifact. A system may influence goals, constraints, and intermediate judgments through suggestions, clarifying questions, or generated alternatives. These indirect effects make human and AI contributions difficult to separate from the output alone.

Recent work has therefore turned to the collaborative process itself. Kim et al. \cite{kim2026cotrace} propose a goal-level attribution framework that decomposes collaborative goals into requirements and traces direct and indirect influence across dialogue turns. Their findings suggest that AI may have limited influence on high-level direction while playing a larger role in shaping lower-level requirements and micro-decisions that users do not always recognize. Other studies show that people assign credit according to the type, amount, and initiative of AI contribution \cite{he2025credit}, while systems such as DraftMarks use process traces to make AI involvement in co-writing more visible \cite{siddiqui2026draftmarks}.

These approaches clarify where AI influence enters collaboration, yet attribution alone cannot resolve accountability. A trace cannot determine whether a contribution was legitimate, whether it should reduce professional credit, or who is responsible when an AI-shaped judgment causes harm. Such questions depend on the obligations and norms of the setting. Contribution traces can provide evidence for accountability decisions, but do not resolve them.

\section{Case 1: Peer Review}

In peer review, reviewers are assigned to assess the contribution of a paper, identify its weaknesses, weigh evidence, and recommend an outcome on behalf of a scholarly community~\cite{lee2013bias}. The review report is only the visible output, thus accountability rests on whether the reviewer understands and can defend the judgments expressed.

Related policies often approach this responsibility through rules about AI use. For example, the ICML 2026 LLM policy compared a strict track that prohibited LLM use in reviewing with a more permissive track that allowed comprehension support and language polishing while limiting the delegation of critical assessment and review writing \cite{icml2026policy}. ICML later reported using hidden prompts to identify possible violations of the strict policy, followed by the desk rejection of hundreds of submissions whose reviews appeared to violate it \cite{icml2026blog}. The episode captures the concern that undisclosed AI use may weaken trust in peer review. It also shows the limits of treating detection as the main response. Detecting model involvement says little about which parts of the review were delegated and whether the reviewer could defend them. These concerns are especially serious given the documented limits of LLMs as paper reviewers and the broader expectation that humans remain responsible for AI-assisted scholarly work \cite{zhou2024reliableReviewer}.

The key distinction is whether the reviewer has taken ownership of the judgment. A model-generated criticism may still be acceptable if the reviewer has independently assessed it, agrees with it, and can defend it from their own understanding of the paper. The problem arises when a reviewer makes an objection, novelty claim, score, or recommendation that they cannot evaluate or explain. A contribution trace can show where a judgment originated or how it was shaped, but it cannot show whether the reviewer has genuinely adopted responsibility for it.

Rules centered on AI use remain useful for setting boundaries and supporting disclosure, but limitations exist as they may penalize acceptable support while remaining vague about the delegation of core evaluative work. Knowing that AI was used does not reveal whether the reviewer understood, endorsed, and could defend the judgments in the review. A fuller account of responsibility therefore requires attention to judgment ownership alongside AI use.

This case points toward accountability infrastructure that distinguishes the roles AI plays and identifies where human ownership must be explicit. In peer review, this could mean separating comprehension support and language editing from critique generation, score calibration, and recommendation. More broadly, institutions should identify the judgments for which a person must understand the reasoning, defend the decision, and take responsibility for the correction. This turns attention from exhaustive records of AI use to the points where accountable judgment matters most.

\section{Case 2: Creative Work in Journalism}

Creative work presents a different accountability problem. In journalism and other communication professions, work is evaluated not only by the quality of the final output, but also by the editorial care, professional judgment, and effort that it reflects \cite{xiao2026collaborative}. AI assistance can therefore change how work is valued even when the worker remains responsible for the result. Disclosure may be read as evidence of reduced care, skill, or authorship. Perceived effort is known to shape judgments of value \cite{kruger2004effort}, while studies of AI-assisted writing have shown related tensions around ownership, authorship, trust, and agency \cite{draxler2024ghostwriter,kadoma2024workplace,jakesch2019aimc}.

A recent study of GenAI use in Chinese newsrooms makes this tension concrete \cite{xiao2026collaborative}. Based on interviews with newsroom managers, editors, and front-line journalists, Xiao et al. found that GenAI was widely used for drafting, summarizing, translation, and language editing, but remained largely private, informal, and disconnected from collaborative workflows. Journalists were often reluctant to discuss these practices with colleagues because AI assistance could be interpreted as laziness, weak editorial rigor, or diminished professional credibility. The result was a gap between what workers did and what they felt safe to acknowledge.

This condition creates what we call a \emph{dark forest of effort}. When AI assistance is useful but disclosure may reduce professional credit, concealment becomes a reasonable response. Workers hide their own processes while seeing only the finished work of others. Organizations then lose the opportunity to develop shared norms around which uses support quality, which delegate editorial judgment, and which should affect credit. Concealment makes it harder to discuss how contributions were produced and where responsibility should lie.

Contribution tracing cannot resolve this tension on its own. Process traces may support coordination and review, but they can also be treated as evidence that a worker performed less ``real'' work. Under such conditions, workers have reason to use private tools, avoid logged systems, or withhold their practices from colleagues. The value of visibility therefore depends on how AI involvement is interpreted and accounted.

This case points toward accountability infrastructure that makes AI involvement safe to discuss. In such a setting, AI assistance should be distinguished from the delegation of editorial judgment, and credit for work grounded in verification, source interpretation, and final responsibility should be preserved. The value of visibility depends on whether workers can disclose AI involvement without a default credit loss. The broader challenge is to make AI contribution speakable before it becomes punishable.

\section{From Attribution to Accountability}

The two cases expose complementary failures. In peer review, a reviewer may remain formally responsible for a judgment they cannot explain or defend, while in creative work, a worker may fully understand and stand behind the final product but lose credit once AI assistance becomes visible. Policies centered on disclosure or detection struggle with both situations, as they say little about how AI involvement should affect responsibility and credit, or who should respond when AI-shaped work goes wrong. Accountability infrastructure thus should make AI-shaped contribution interpretable and discussable before it becomes grounds for penalty.

A practical starting point is to \textbf{describe the role of AI} beyond recording the presence. Editing language, making explanations, proposing alternatives, planning a workflow, verifying a claim, and recommending a decision affect collaborative work in different ways. Some roles mainly alter expression, while others shape evaluation, direction, or action. These differences matter because the relevant question is how the model participated in the judgments on which the work depends. Research on contribution attribution similarly shows that credit judgments vary with the type and initiative of AI contribution \cite{he2025credit}. Disclosure should preserve these distinctions instead of reducing them to a single category of AI use.

In addition, \textbf{judgment ownership checkpoints} are needed. Institutions can identify decisions for which a person must be able to explain the reasoning, endorse the outcome, respond to challenge, and take responsibility for correction. In peer review, these points can include novelty assessment, weakness identification, and the final recommendation. In journalism, they may include factual framing, source interpretation, editorial voice, and publication risk. Such checkpoints do not require a complete record of every prompt or revision. The purpose is to identify moments where someone must be able to say: I understand this judgment, I endorse it, and I am responsible for its consequences.

A further requirement is to \textbf{protect disclosure and enable repair}. Workers have little reason to describe AI involvement honestly when disclosure reduces credit by default or creates a presumption of misconduct. Accountability infrastructure therefore needs shared expectations about acceptable AI roles, channels through which assistance can be disclosed without immediate sanction, and procedures for challenging and repairing AI-shaped work. Accountability extends far beyond the assignment of blame, which concerns who must explain an error, revise a decision, and address the potential error. Without these protections, disclosure may reproduce the tensions around ownership, trust, and agency already documented in AI-assisted writing \cite{draxler2024ghostwriter,jakesch2019aimc,hohenstein2020moral}.

This provocation does not assume that more visibility is always desirable. Detailed process records can clarify how AI shaped a collaborative outcome, but they may also enable surveillance or simplistic judgments about effort. Attribution remains valuable because it reveals forms of influence that final artifacts cannot show. However, its value depends on the norms through which institutions interpret that evidence. Provenance can establish that AI participated in the process, but the accountability infrastructure must address more complicated questions, including whether that participation is acceptable, how it should affect credit, whether a person owns the resulting judgment, and what should happen when the work is challenged or causes errors.

\section{Conclusion}

Contribution dissolution is becoming a normal condition of human-AI collaboration. Disclosure and provenance can reveal AI involvement, but cannot determine who understands, endorses, and answers for the resulting judgments. The two cases expose the cost of this gap: responsibility without ownership in peer review and ownership without credit in creative work. Accountability should therefore focus on the roles AI plays, the judgments that require clear human ownership, and the conditions under which AI involvement can be disclosed, challenged, and corrected.

\bibliographystyle{ACM-Reference-Format}
\bibliography{reference}

\end{document}